\documentclass[letterpaper]{article} 
\usepackage{aaai25}  
\usepackage{times}  
\usepackage{helvet}  
\usepackage{courier}  
\usepackage[hyphens]{url}  
\usepackage{graphicx} 
\usepackage{natbib}  
\usepackage{caption} 
\usepackage{algorithm}
\usepackage{algorithmic}
\usepackage{multirow}
\usepackage{colortbl}
\PassOptionsToPackage{table,xcdraw}{xcolor}
\usepackage[table,xcdraw]{xcolor}
\usepackage{makecell}
\usepackage{amssymb}
\usepackage{amsmath}
\nocopyright
\usepackage{newfloat}
\usepackage{listings}
\DeclareCaptionStyle{ruled}{labelfont=normalfont,labelsep=colon,strut=off} 
\floatstyle{ruled}
\newfloat{listing}{tb}{lst}{}
\floatname{listing}{Listing}
\title{Adaptive Dual-domain Learning for Underwater Image Enhancement}
\author{
    Lintao Peng \textsuperscript{\rm 1}, Liheng Bian \textsuperscript{\rm 1} \thanks{Corresponding author.}\\
}
\affiliations{
    \textsuperscript{\rm 1}Beijing Institute of Technology, Beijing \& Zhuhai, China\\

    \{penglintao, bian\}@bit.edu.cn
}

\usepackage{bibentry}

\begin{document}

\maketitle

\begin{abstract}
Recently, learning-based Underwater Image Enhancement (UIE) methods have demonstrated promising performance. However, existing learning-based methods still face two challenges. 1) They rarely consider the inconsistent degradation levels in different spatial regions and spectral bands simultaneously. 2) They treat all regions equally, ignoring that the regions with high-frequency details are more difficult to reconstruct. To address these challenges, we propose a novel UIE method based on spatial-spectral dual-domain adaptive learning, termed SS-UIE. Specifically, we first introduce a spatial-wise Multi-scale Cycle Selective Scan (MCSS) module and a Spectral-Wise Self-Attention (SWSA) module, both with linear complexity, and combine them in parallel to form a basic Spatial-Spectral block (SS-block). Benefiting from the global receptive field of MCSS and SWSA, SS-block can effectively model the degradation levels of different spatial regions and spectral bands, thereby enabling degradation level-based dual-domain adaptive UIE. By stacking multiple SS-blocks, we build our SS-UIE network. Additionally, a Frequency-Wise Loss (FWL) is introduced to narrow the frequency-wise discrepancy and reinforce the model's attention on the regions with high-frequency details. Extensive experiments validate that the SS-UIE technique outperforms state-of-the-art UIE methods while requiring cheaper computational and memory costs.
\end{abstract}
\begin{links}
  \link{Code}{https://github.com/LintaoPeng/SS-UIE}
\end{links}

%

\section{Introduction}

\begin{figure}[ht] 
	\centering
	\includegraphics[width=0.95\linewidth]{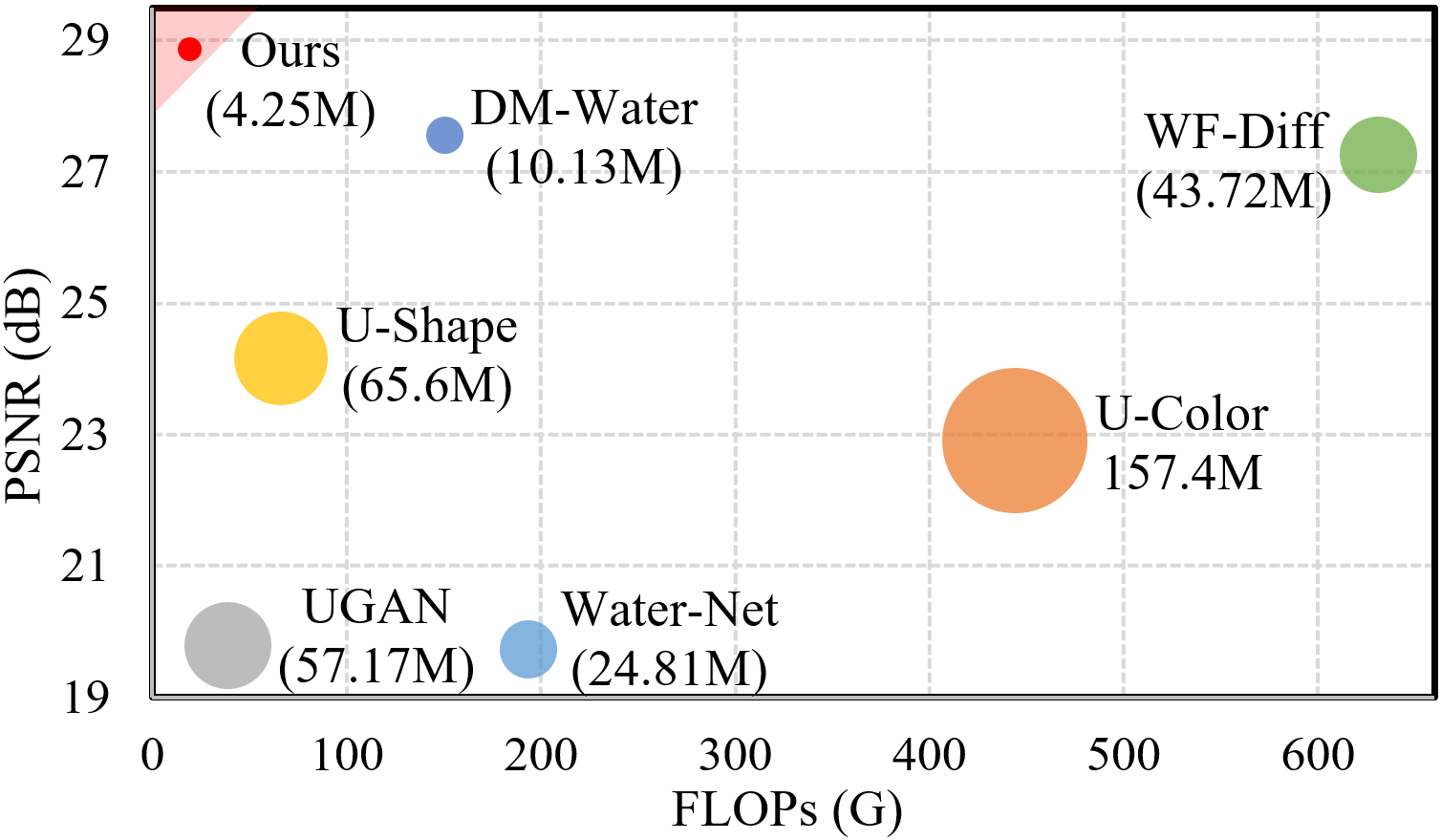}
	\caption{PSNR-Parameters-FLOPs comparisons with existing UIE methods. The vert axis is PSNR (dB), the horizontal axis is FLOPs (computational cost), and the circle radius is the parameter number (memory cost) of each UIE model. Our SS-UIE outperforms state-of-the-art (SOTA) methods while requiring fewer FLOPs and Params. 
	}
	\label{fig1}
\end{figure}

\label{sec:intro}
Underwater imaging technology \cite{sahu2014survey,yang2019depth} is essential for obtaining underwater images and investigating the underwater environment. However, due to the strong absorption and scattering effects on light caused by dissolved impurities and suspended matter in the medium (water), underwater images often suffer from inconsistent levels of degradation in different spatial regions and spectral bands, which ultimately leads to problematic issues, such as color casts, artifacts and blurred details \cite{schettini2010underwater}. Therefore, Underwater Image Enhancement (UIE) related innovations are of great significance for improving the visual quality and merit of images in accurately understanding the underwater world.

The purpose of UIE is to enhance image quality by removing scattering and correcting color distortions in degraded images. To achieve this goal, traditional physical model-based methods \cite{Ancuti2012fusion,Drews2013UDCP,Li2016miniinfor} first investigate the physical mechanisms of underwater image degradation and then reverse the degradation process using various hand-crafted physical priors. However, these methods rely on manually tuned physical parameters and cannot account for all the complex physical and optical factors in underwater scenes, leading to poor generalizability. In contrast, learning-based UIE methods eliminate the need for manual parameter tuning. Researchers initially applied CNNs \cite{zeiler2014visualizing} to learning-based UIE methods to improve generalization capability. However, CNN-based methods \cite{li2017watergan,Li2019UIEB,Li2021UnderwaterIE,Islam2020FUnIE} show limitations in capturing non-local self-similarity and long-range dependencies in images, resulting in suboptimal UIE performance.

Recently, Transformer-based UIE methods \cite{Peng2023ushape,tang2023underwater,zhao2024wavelet} have demonstrated promising performance. The self-attention mechanism in Transformer \cite{vaswani2017attention} can model long-range dependencies and non-local similarities, which offers the possibility to address the shortcomings of CNN-based methods. However, the existing Transformer-based methods still have the following issues.
\textbf{First}, in the global attention Transformer \cite{dosovitskiy2020image}, the computational complexity is quadratic to the spatial size, this burden is non-trivial and sometimes unaffordable. \textbf{Second}, existing Transformer-based UIE methods ignore the physical prior that underwater images have inconsistent degradation levels in different spatial regions and spectral bands \cite{sahu2014survey,yang2019depth}, and still treat all spatial regions and spectral bands equally, resulting in unsatisfactory enhancement results.

Naturally, a question arises: \textit{How to achieve spatial-spectral dual domain adaptive UIE with linear complexity?} The recently proposed state space model, i.e., Mamba \cite{gu2023mamba} inspired us, as it can obtain global receptive field and content-based reasoning ability with linear complexity \cite{han2024demystify}. Based on this, we propose a spatial-wise Multi-scale Cycle Selective Scan (MCSS) module to obtain the global receptive field with linear complexity, thereby modeling the degradation levels in different spatial regions.
We then combine it with the proposed Spectral-Wise Self-Attention (SWSA) module in parallel to create a basic Spatial-Spectral block (SS-block). SWSA uses a fast FFT \cite{nussbaumer1982fast} to transform the image into the frequency domain and applies a learnable global filter to mix frequency-wise tokens. This frequency-domain processing method can create global correlations for each spectral band and capture the inter-spectral long-distance dependencies, thereby modeling the degradation levels in different spectral bands.
Thanks to the parallel design, SS-block can model the degradation levels in different spatial regions and spectral bands simultaneously with linear complexity, thereby achieving degradation level-based dual-domain adaptive UIE. By stacking multiple SS-blocks, we build our spatial-spectral dual-domain adaptive learning UIE network, termed SS-UIE.
In addition, considering that the spatial regions with high-frequency details are more difficult to reconstruct, we introduce a Frequency-Wise Loss (FWL) function to reinforce the network’s attention in the regions with rich detail. Equipped with the proposed techniques, our SS-UIE achieves SOTA UIE performance, as shown in Fig. \ref{fig1}.
Our contributions can be summarized as follows,
\begin{itemize}
	\item  Our proposed MCSS and SWSA module can obtain the spatial-wise and spectral-wise global receptive fields with linear complexity, respectively, thereby modeling the degradation levels in different spatial regions and spectral bands.
	
	\item We combined MCSS and SWSA in parallel to form an SS-block, which can reinforce the network's attention to the spatial regions and spectral bands with serious attenuation, and achieve degradation level-based adaptive UIE.  
	
	\item The proposed FWL function can narrow the frequency-wise discrepancy, and force the model to restore high-frequency details adaptively without additional memory and computational costs.
	
\end{itemize}

\section{Related Work}

\subsection{Underwater Image Enhancemnet}

\textbf{Physical model-based methods.} 
Traditional physical model-based UIE methods address underwater image degradation by employing hand-crafted priors and estimated parameters. Prior is the basis of the physical model-based UIE, in which existing work includes underwater dark channel priors \cite{Drews2013UDCP}, attenuation curve priors \cite{Wang2018aacp}, fuzzy priors \cite{Chiang2011WCD} and minimum information priors \cite{Li2016miniop}, etc. For example, Akkaynak et. al \shortcite{Akkaynak2019seathru} proposed a method based on the revised physical imaging model. However, the depth map of the underwater image is difficult to obtain, which leads to unstable performance.  Consequently, hand-crafted priors limit the model’s robustness and scalability in complex scenarios.

\textbf{CNN-based methods.} Existing CNN-based UIE methods can be divided into two main technical routes, (1) \emph{designing an end-to-end UIE model;} (2) \emph{utilizing deep-learning models to estimate physical parameters, and then restoring the clean image based on the degradation model.} To alleviate the need for real-world underwater paired training data, many end-to-end methods introduced GAN-based \cite{goodfellow2020generative} framework for UIE, such as Water GAN \cite{li2017watergan}, UGAN \cite{Fabbri2018UGAN} and UIE-DAL \cite{uplavikar2019UIEDAL}. However, the training datasets used by the above methods are not matched real underwater images, which leads to limited enhancement effects in diverse real-world underwater scenes. Recently, Li et al. \shortcite{Li2021UnderwaterIE} combined the underwater physical imaging model and designed a medium transmission guided model to reinforce the network’s response to areas with more severe quality degradation.  Nevertheless, physical models sometimes fail with varied underwater environments. Compared with physical model-based methods, CNN-based methods have improved generality and speed.  However, CNN-based methods still have limitations in capturing long-distance dependencies and non-local similarities.

\textbf{Transformer-based methods.} Recently, the Global Vision Transformer has achieved great success in image classification \cite{dosovitskiy2020image}. However, for dense image processing tasks such as UIE, the Transformer's computational complexity scales quadratically with image size, this burden is non-trivial and sometimes unaffordable. Local-window self-attention \cite{liu2021swin} can effectively reduce the computational complexity, but its receptive field is quite limited. Moreover, existing Transformer-based UIE methods \cite{Peng2023ushape,tang2023underwater,10484001} often ignore the inconsistent degradation level of underwater images in different spatial regions and spectral bands, and still treat all spatial regions and spectral bands equally, resulting in unsatisfactory enhancement results. In this work, we introduce the MCSS and SWSA modules with linear complexity and combine them in parallel to form the SS-block, which can simultaneously model the degradation levels of different spatial regions and spectral bands, and achieve degradation level-based adaptive UIE.

\subsection{State Space Models}

Originating from control theory, State Space Models (SSMs)\cite{gu2021efficiently} have garnered increasing attention due to their efficacy in long-term language modeling. Unlike self-attention based transformers, most SSMs capture long-range token interactions through linear recurrent processes, entailing $\mathcal{O}(N)$ complexity theoretically. Mamba \cite{gu2023mamba,xu2024survey} improves the expressiveness of SSMs by introducing a selective mechanism, with its structural parameters adaptively learned from inputs. Motivated by its potential for modeling high-resolution images, many researchers try to apply Mamba to vision tasks. For instance, VMamba \cite{liu2024vmamba} introduces a cross-scan module to enable 1D selective scanning in 2D image space. LocalMamba \cite{huang2024localmamba} utilizes local windows to enhance local modeling capability. EfficientVMamba \cite{pei2024efficientvmamba} designs an atrous-based selective scan approach to enhance efficiency. In this work, 
inspired by the physical prior that underwater images have different degradation levels in different spatial regions, we designed the MCSS module based on Mamba. Its global receptive field enables it to model the degradation levels in different regions and reinforce the network‘s attention on severely degraded regions, achieving adaptive UIE.

\section{Method}

\subsection{The Overall Architecture of SS-UIE Network}

\begin{figure*}[h] 
	\centering
	\includegraphics[width=0.95\linewidth]{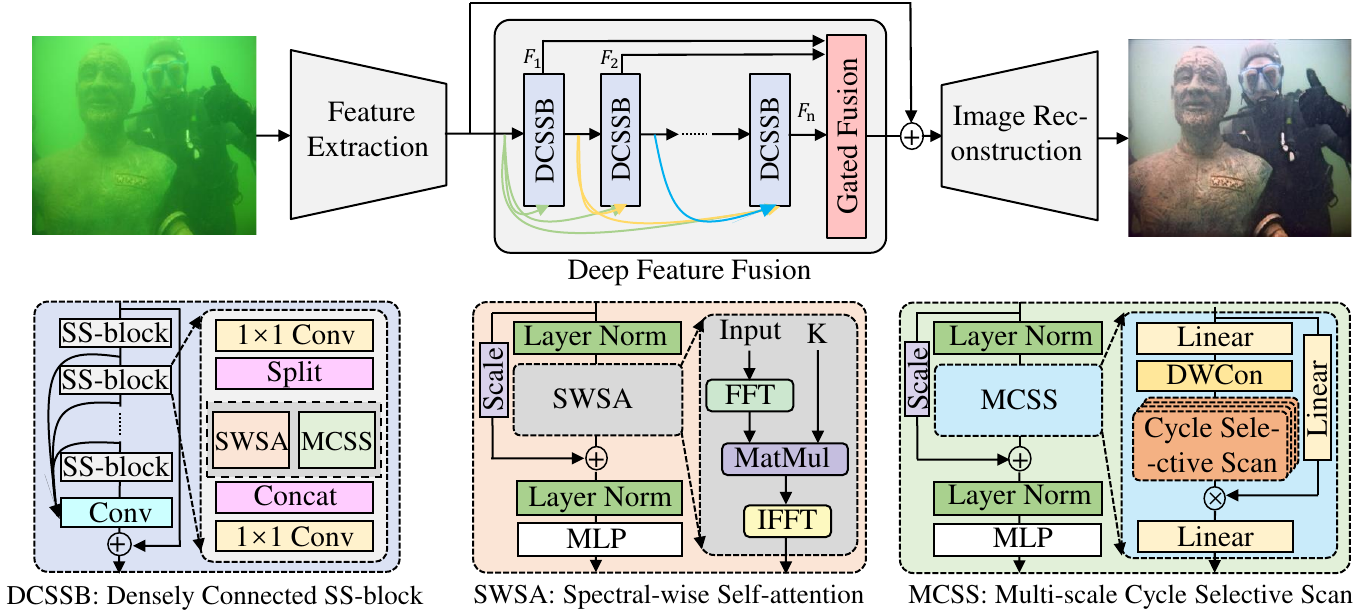}
	\caption{The overall structure of SS-UIE. We combine the spatial-wise Multi-scale Cycle Selective Scan (MCSS) module with the Spectral-Wise Self-Attention (FWSA) module in parallel to form the Spatial-Spectral block (SS-block). MCSS and SWSA can obtain the spatial-wise and spectral-wise global receptive field with linear complexity, respectively. The parallel design facilitates complementary interactions between spatial and spectral features, enabling the SS-block to capture the degradation levels in different spatial regions and spectral bands, thereby achieving degradation level-based dual-domain adaptive UIE.}
	\label{method}
\end{figure*}

We designed the SS-UIE network based on the SS-block. As shown in Fig. \ref{method}, the network consists of three modules, including the shallow feature extraction module, the deep feature fusion module, and the image reconstruction module.

\textbf{Shallow Feature Extraction}. Given a low-quality underwater image $\mathbf{I}_{LQ}\in\mathbb{R}^{H\ast W\ast C}$ ($H$, $W$ and $C$ are the image’s height, width and channel number), we use the shallow feature extractor $\rm H_{FE}\left(\ast\right)$ to explore its shallow features $\mathbf{F}_0\in\mathbb{R}^{\frac{H}{2^{i}}*\frac{W}{2^{i}}*2^{i}C}$ as, 
\begin{equation}
	\begin{aligned}
		\mathbf{F}_{0}={\rm H_{FE}}(\mathbf{I}_{LQ}),
	\end{aligned}
	\label{eq:5}
\end{equation}
where $i$ represents the number of downsampling convolution blocks. After each convolution block, the resolution of feature map $\mathbf{F}$ is halved and the number of channels is doubled.

\textbf{Deep Feature Fusion}. Next, we use 6 Densely Connected SS-Blocks (DCSSB) to extract different levels of high dimensional features $\mathbf{F}_j\in\mathbb{R}^{\frac{H}{2^{i}}*\frac{W}{2^{i}}*2^{i}C}$ from $\mathbf{F}_0$ to $\mathbf{F}_{j-1}$, denoted as,
\begin{equation}
	\begin{aligned}
		\mathbf{F}_j={\rm H_{DCSSB}^{j}}\left(\mathbf{F}_0,\ \mathbf{F}_1,\ldots,\ \mathbf{F}_{j-1}\right),\ 
	\end{aligned}
	\label{eq:6}
\end{equation}
where $\rm \mathbf{H}_{DCSSB}^{j}$ represents the $j_{th}$ DCSSB block, which consists of 4 SS-blocks and a convolution layer. 
The last block of the deep feature fusion module is the gated fusion block, which fuses the outputs of different DCSSB blocks with self-adaptive weights $w_i$, 
\begin{equation}
	\begin{aligned}
		\mathbf{F}_{DF}=w_1\mathbf{F}_1+w_2\mathbf{F}_2+\ldots+w_n\mathbf{F}_n,\ 
	\end{aligned}
	\label{eq:7}
\end{equation}
where $\mathbf{F}_{DF}$ represents the multi-level deep fusion features output by the gated fusion block. The self-adaptive weights $w_i$ control the weighted fusion of each DCSSB's output, which are adaptively adjusted through backpropagation during network training. Such a module structure is conducive to deep mining of different levels of high dimensional features, which prevents losing long-term memory as the network deepens and enhances local details.

\textbf{Image Reconstruction}. The image reconstruction module consists of several upsampling convolutional blocks. We retrieve high-quality underwater images by aggregating shallow features and multi-level deep fusion features. The operation is described as,
\begin{equation}
	\begin{aligned}
		\mathbf{I}_{pred}={\rm H_{REC}}\left(\mathbf{F}_0+\mathbf{F}_{DF}\right).\ 
	\end{aligned}
	\label{eq:8}
\end{equation}

The shallow features $\mathbf{F}_0$ are mainly low-frequency image features, while the multi-level deep fusion features $\mathbf{F}_{DF}$ focus on recovering lost medium-frequency and high-frequency features.

\subsection{Spatial-Spectral Block}
The detailed structure of SS-block is shown in Fig. \ref{method}. Assuming that the inputs of SS-block are feature maps $\mathbf{X}_{in}\in\mathbb{R}^{\frac{H}{2^{i}}*\frac{W}{2^{i}}*2^{i}C}$ at different scales.  To be specific, for an input feature map $\mathbf{X}_{in}$, it is first passed through a $1\times 1$ convolution and split evenly into two feature maps $\mathbf{X}_1$ and $\mathbf{X}_2$,

Next,   $\mathbf{X}_1\in\mathbb{R}^{\frac{H}{2^{i}}*\frac{W}{2^{i}}*2^{i}C}$ is fed into the SWSA module for further processing. As shown in Fig. \ref{method}, in the SWSA, we propose a global filter $\mathbf{K}$ as an alternative to the self-attention layer which can mix tokens representing different spectrum information. For input feature $\mathbf{X}_1$, we first perform layer normalization (LN). Then, we use 2D FFT  to convert $\mathbf{X}_1$ to the frequency domain,
\begin{equation}
	\mathbf{X}_{F}=\mathcal{F}({\rm LN}(\mathbf{X}_{1})),
	\label{eq:10}
\end{equation}
where $\mathcal{F}(\cdot)$ denotes the 2D FFT. The output feature $\mathbf{X}_{F}$ represents the spectrum of $\mathbf{X}_{1}$. We can then modulate the spectrum by multiplying a learnable filter $ \mathbf{K}\in\mathbb{R}^{\frac{H}{2^i} *\frac{W}{2^i} *2^iC}$ to the  $\mathbf{X}_{F}$,
\begin{equation}
	\mathbf{Y}_F=\mathbf{K}\odot \mathbf{X}_F,
	\label{eq:11}
\end{equation}
where $\odot $ is the element-wise multiplication (Hadamard product). The filter K has the same dimension with $\mathbf{X}_F$, which can represent an arbitrary filter in the spectral domain and enables global spectral information exchange in the SWSA module. Finally, we adopt the IFFT to transform the modulated spectrum $\mathbf{Y}_F$ back to the spatial domain and add with a residual value controlled by the learnable scale factor $\alpha_1$ to get the $\mathbf{Y}_{1}$,
\begin{equation}
	\mathbf{Y}_{1}={\rm MLP}({\rm LN}(\mathcal{F}^{-1}(\mathbf{Y}_{F})+\alpha_1\ast \mathbf{X}_1)),
	\label{eq:12}
\end{equation}
where $\mathcal{F}^{-1}(\cdot)$ denotes the 2D IFFT.
The global filter $\mathbf{K}$ is motivated by the frequency filters in the digital image processing \cite{pitas2000digital}, which can be regarded as a set of learnable frequency filters for different hidden dimensions. 

\begin{figure*}[h] 
	\centering
	\includegraphics[width=1.0\linewidth]{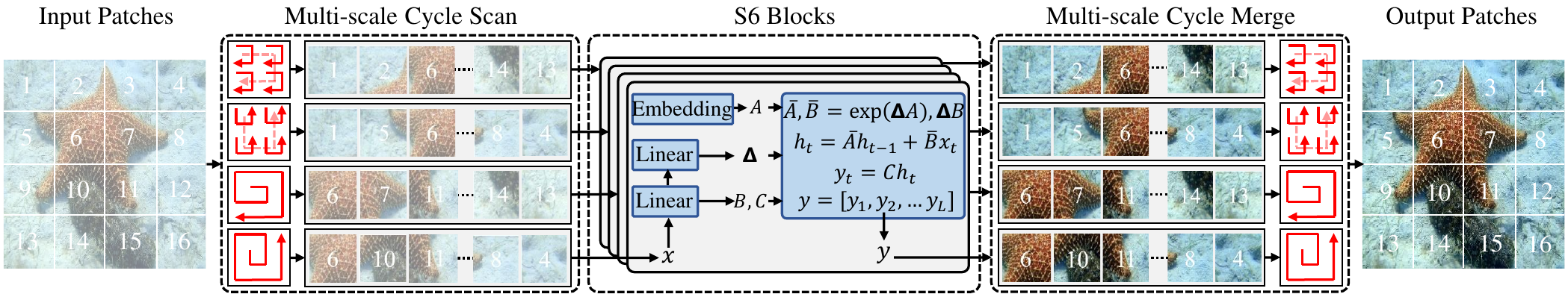}
	\caption{The framework of MCSS module. Given the input data $\mathbf{X}$, MCSS first unfolds input patches into sequences along multiple distinct traversal paths (i.e., multi-scale cycle scan), processes each patch sequence using a separate S6 block in parallel, and subsequently reshapes and merges the resultant sequences to form the output map (i.e., feature-merge).} 
	\label{ss-block}
\end{figure*}

Similarly, $\mathbf{X}_2\in\mathbb{R}^{\frac{H}{2^{i}}*\frac{W}{2^{i}}*2^{i}C}$ is sent to the MCSS module. Our MCSS module is built on the Select-Scan Structured State Space for Sequences (S6) block \cite{gu2023mamba}. The S6 block is the core of Mamba and can achieve global receptive field, dynamic weight, and linear complexity at the same time. While the sequential nature of the scanning operation in S6 aligns well with NLP tasks involving temporal data, it poses a significant challenge when applied to vision data, which is inherently nonsequential and encompasses spatial information (e.g., local texture and global structure). In order to simultaneously perceive local textures and global features in visual data, we propose the multi-scale cycle selective-scan strategy (as shown in Fig. \ref{ss-block}) to adapt S6 to vision data without compromising its advantages.

As shown in Fig. \ref{method}, the input feature $\mathbf{X}_2$ will go through two parallel branches. In the first branch, the feature's channel number is expanded to $\lambda{2^{i}C}$ by a linear layer, where $\lambda$ is a pre-defined channel expansion factor. Subsequently, the feature is processed by a depth-wise convolution (DWConv) \cite{han2023connection} and an MCSS layer (as shown in Fig. \ref{ss-block}).
In the second branch, the features channel is also expanded to $\lambda{2^{i}C}$ with a linear layer. The linear layer consists of a $1\times1$ convolution and a SiLu activation function. After that, features from the two branches are aggregated with the Hadamard product. Finally, the channel number is projected back to $2^{i}C$ to generate output $\mathbf{Y}_2^{'}$ with the same shape as input:
\begin{equation}
	\begin{aligned}
		\mathbf{X}_2^{'}&={\rm MCSS}({\rm DWConv}({\rm Linear}({\rm LN}(\mathbf{X}_2)))),\\
		\mathbf{Y}_2^{'}&={\rm Linear}(\mathbf{X}_2^{'}\odot({\rm Linear}(\mathbf{X}_2))).
	\end{aligned}
	\label{eq:13}
\end{equation}

Then, $\mathbf{Y}_2^{'}$ needs to added with a residual value controlled by the learnable scale factor $\alpha_2$ to get the output of MCSS module,
\begin{equation}
	\begin{aligned}
		\mathbf{Y}_2={\rm MLP}({\rm LN}(\mathbf{Y}_2^{'}+\alpha_2\ast \mathbf{X}_2)).
	\end{aligned}
	\label{eq:14}
\end{equation}

Finally, $\mathbf{Y}_1$ and $\mathbf{Y}_2$ are concatenated as the input of a $1\times1$ convolution, which has a residual connection with the input $\mathbf{X}_{in}$. As such, the final output of SS-block is given by,
\begin{equation}
	\begin{aligned}
		\mathbf{X}_{out}={\rm Conv1}\times1({\rm Concat}(\mathbf{Y}_1,\mathbf{Y}_2))+\mathbf{X}_{in}.
	\end{aligned}
	\label{eq:15}
\end{equation}

\subsection{Frequency-wise Loss}

The F-Principle \cite{xu2019training} proves that deep learning networks tend to prefer low frequencies to fit the objective, which will result in the frequency domain gap. Recent studies \cite{wang2019cnngenerated,9035107} indicate that the periodic pattern shown in the frequency spectrum may be consistent with the artifacts in the spatial domain. In UIE task, the model overfitting at low frequencies brings smooth textures and blurry structures. So exploring adaptive constraints on specific frequencies is essential for the refined reconstruction.

\begin{table*}[h]
	\centering
	\renewcommand\arraystretch{1.2}
	\begin{tabular}{c|c|cccccccc|c}
		\Xhline{1.5px}
		\rowcolor[HTML]{EFEFEF} 
		& Methods & Fusion & MLLE  & Water-Net & UGAN & U-Color & U-Shape & DM-Water &WF-Diff & Ours \\ \hline
		\multirow{4}{*}{UIEB} & FID↓ &38.28 &37.89 & 37.48  &42.98  & 38.25   & 46.11   & 31.07  &\underline{27.85}      &  \textbf{27.31}    \\
		& LPIPS↓  & 0.239  & 0.217 & 0.212     & 0.204 & 0.234   & 0.226   & 0.144    &\underline{0.125}   &  \textbf{0.117}    \\
		& PSNR↑   & 19.04  & 20.47 & 19.35     & 20.68 & 20.71   & 21.25   & 21.88    &\underline{23.86}      &  \textbf{24.19}    \\
		& SSIM↑   & 0.821  & 0.862 & 0.832     & 0.842 & 0.841   & 0.845   & 0.819    &\underline{0.873}   &  \textbf{0.897}   \\ \hline
		\multirow{4}{*}{LSUI} & FID↓ &39.25 & 30.78 & 38.90 & 32.66  & 45.06   & 28.56   & 27.91      &\underline{26.75}   &  \textbf{25.98}    \\
		& LPIPS↓  & 0.244  & 0.153 & 0.168     & 0.203 & 0.123   & \textbf{0.103}   & 0.114    &0.110      &  \underline{0.106}   \\
		& PSNR↑   & 17.48  & 21.55 & 19.73     & 19.79 & 22.91   & 24.16   & \underline{27.65}    &27.26      &  \textbf{28.87}    \\
		& SSIM↑   & 0.791  & 0.892 & 0.823     & 0.782 & 0.891   & 0.932   & 0.887    &\underline{0.944}      &  \textbf{0.952}    \\ \hline
		\multirow{2}{*}{U45}  & UIQM↑  & 2.623   & 2.683  &2.951    & 3.033  & 3.104   & 3.151   & 3.086 &\underline{3.181}  &\textbf{3.246}      \\
		& UCIQE↑  & 0.612   & 0.456  & 0.601     & 0.573  & 0.586   & 0.592   & \underline{0.634}   &0.619      &\textbf{0.703}    \\ \Xhline{1.5px}
	\end{tabular}
	\caption{Quantitative comparison of different UIE methods on the UIEB \cite{Li2019UIEB}, LSUI \cite{Peng2023ushape} and U45 \cite{li2019fusion} datasets. The best results are highlighted in bold and the second best results are underlined.}
	\label{tab1}
\end{table*}
\begin{figure*}[h] 
	\centering
	\includegraphics[width=1.0\linewidth]{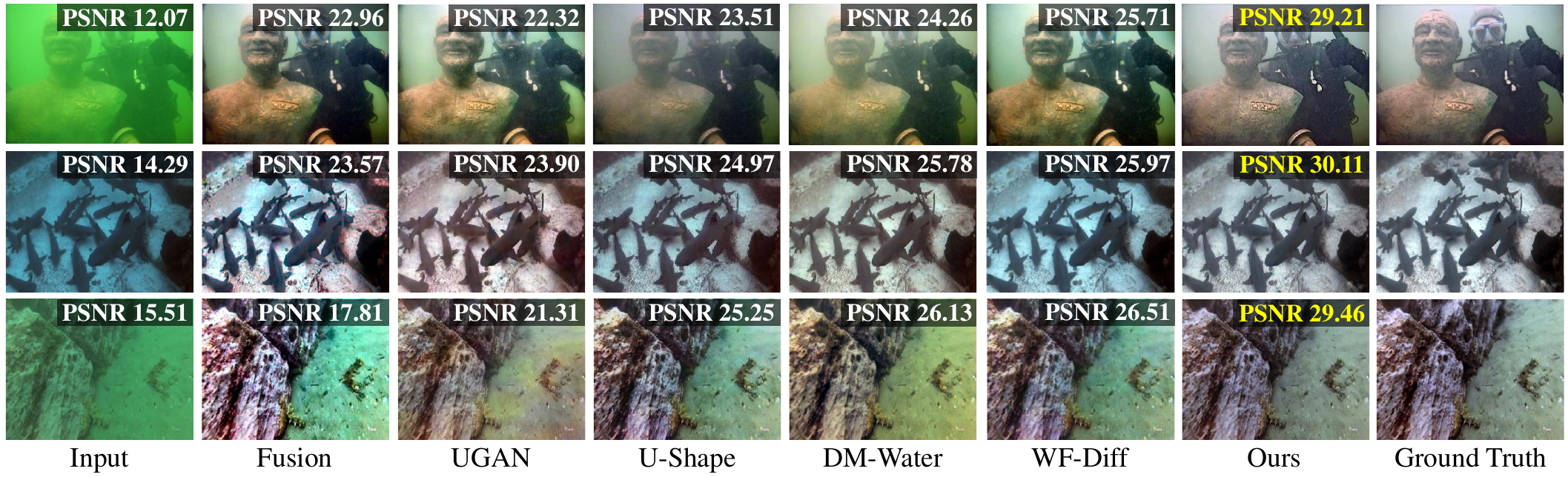}
	\caption{Visual comparison of enhancement results sampled from the test set of LSUI and UIEB dataset. The highest PSNR scores are marked in yellow. It can be seen that the enhancement results of our method are the closest to the ground truth. } 
	\label{full}
\end{figure*}

In this work, we use the Discrete Fourier Transform (DFT) to convert the underwater images to the frequency domain and to supervise the frequency distance between truth and predicted images adaptively. We calculate the frequency spectrum for each channel. In a specific channel $k$, the conversion relationship between spatial coordinates $(h, w, k)$ and frequency domain coordinates $(u, v)$ is expressed as,
\begin{equation}
	\begin{aligned}
		\mathbf{F}^{k}(u,v)&=\sum_{h=0}^{H-1}\sum_{w=0}^{W-1}\mathbf{I}(h,w,k)e^{-j2\pi(\frac{uh}{H}+\frac{vw}{W})},
	\end{aligned}
	\label{eq:16}
\end{equation}
where $\mathbf{F}$ is the frequency spectra of all channels corresponding to the underwater image $\mathbf{I}$.
Then we can calculate the dynamic weights based on the frequency distance to make the model concentrate on high-frequency details that are difficult to reconstruct.
Specifically, In each channel $k$, the frequency distance between ground truth and predicted image is equivalent to the power distance between their spectrum $\mathbf{F}_{gt}^k$ and $\mathbf{F}_{pred}^k$, which is defined as,
\begin{equation}
	\begin{aligned}
		d^k(u,v)=\left\|\mathbf{F}_{gt}^k(u,v)-\mathbf{F}_{pred}^k(u,v)\right\|^{2}. 
	\end{aligned}
	\label{eq:17}
\end{equation}

Next, we define a dynamic weight factor $\theta(u,v)$ linearly related to the distance $d(u,v)$ to make the model pay more attention to the frequencies that are hard to synthesize. Then the distance between the ground truth and the predicted image in a single channel $k$ is formulated as,
\begin{equation}
	\begin{aligned}
		d(\mathbf{F}_{gt}^{k},\mathbf{F}_{pred}^{k})=\frac{1}{HW}\sum_{u=0}^{H-1}\sum_{v=0}^{W-1}\theta^{k}(u,v)d^{k}(u,v),
	\end{aligned}
	\label{eq:18}
\end{equation}
where the $\theta^{k}(u,v)$ changes linearly with the absolute value of the $k_{th}$ channel frequency distance $\sqrt{(|d^k(u,v)|)}$. We traverse $k = \{0,1,2...C-1\}$ and sum each distance to calculate the frequency domain loss $ {\rm L}_{FWL}$ as,
\begin{equation}
	\begin{aligned}
		{\rm L}_{FWL}(\mathbf{F}_{gt},\mathbf{F}_{pred})=\sum_{k=0}^{C-1}d(\mathbf{F}_{gt}^k,\mathbf{F}_{pred}^k).
	\end{aligned}
	\label{eq:19}
\end{equation}

In addition, we also used the spatial-wise ${\rm L}_1$ loss, and combined it with the $ {\rm L}_{FWL}$ to form a dual domian loss,
\begin{equation}
	\begin{aligned}
		{\rm L}_{total}={\rm L}_{1}(\mathbf{I}_{gt},\mathbf{I}_{pred})+\lambda {\rm L}_{FWL}(\mathbf{F}_{gt},\mathbf{F}_{pred}),
	\end{aligned}
	\label{eq:20}
\end{equation}
where $\lambda$ is a weight factor, which makes the values of ${\rm L}_{1}$ and ${\rm L}_{FWL}$ be of the same order of magnitude.

\section{Experiments}
\subsection{Experiment Setup}

\textbf{Datasets.} We utilize the datasets UIEB \cite{Li2019UIEB} and LSUI \cite{Peng2023ushape} to evaluate our model. The UIEB dataset comprises 890 images with corresponding labels. Out of these, 800 images are allocated for training, and the remaining 90 are designated for testing. The LSUI dataset is randomly partitioned into 3879 images for training and 400 images for testing. To ensure fairness, we only divide the LSUI dataset once, and then use it for the training and testing of all UIE methods involved in the comparison.
In addition, to verify the generalization of SS-UIE, we use non-reference benchmarks U45 \cite{li2019fusion}, which contains 45 underwater images for testing.

\textbf{Evaluation Metrics.} For the full-reference experiment, we primarily utilize two metrics: PSNR and SSIM \cite{1284395}. Higher PSNR and SSIM values signify superior quality of the generated images. Additionally, we incorporate the LPIPS and FID metrics for full-reference image evaluation. LPIPS \cite{8578166} is a deep neural network-based image quality metric that assesses the perceptual similarity between two images. FID \cite{fid} measures the distance between the distributions of two images. A lower LPIPS and FID score indicates a more effective UIE approach. For non-reference benchmark U45, we introduce UIQM \cite{Panetta2015UIQM} and UCIQE \cite{Yang2015UCIQE} to evaluate our method.

\textbf{Comparison Methods.} We conduct a comparative analysis between our SS-UIE and eight state-of-the-art (SOTA) UIE methods, namely Fusion \cite{Ancuti2012fusion}, MLLE \cite{mlle}, Water-Net \cite{Li2019UIEB}, UGAN \cite{Fabbri2018UGAN}, U-Color \cite{Li2021UnderwaterIE}, U-Shape \cite{Peng2023ushape}, DM-Water \cite{tang2023underwater}, and WF-Diff \cite{zhao2024wavelet}. 
To ensure a fair comparison, we utilize the provided source codes from the respective authors and adhere strictly to the identical experimental settings across all evaluations.

\begin{figure*}[h] 
	\centering
	\includegraphics[width=1.0\linewidth]{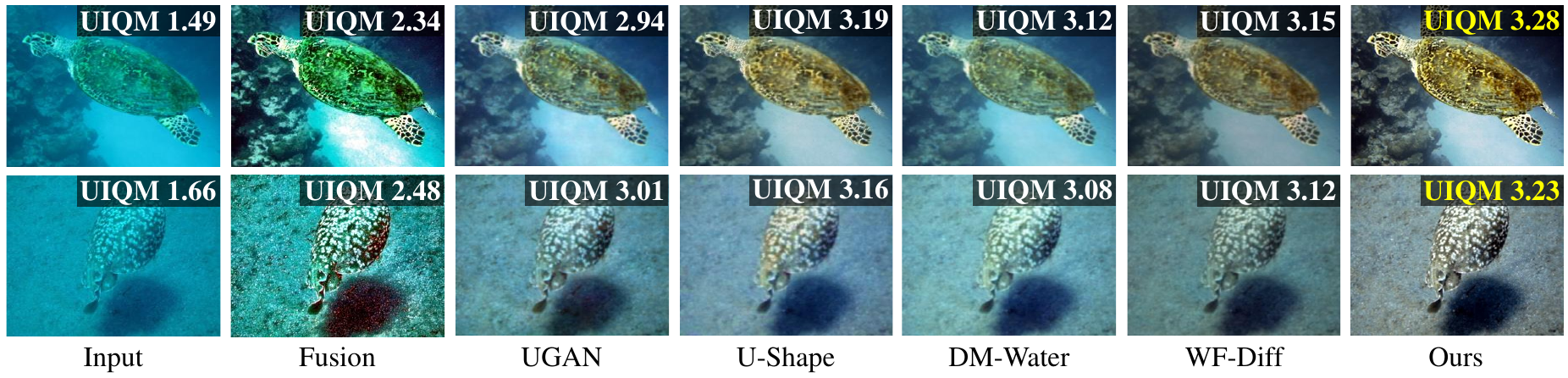}
	\caption{Visual comparison of the non-reference evaluation sampled from the U45 dataset. Compared with existing methods, our method exhibits fewer color casts and artifacts, and recovers high-frequency local details better.} 
	\label{non}
\end{figure*}


\subsection{Experiment Results}

\textbf{Full-Reference Evaluation.} The test sets of UIEB and LSUI datasets were used for evaluation. The statistical results and visual comparisons are presented in Tab. \ref{tab1} and Fig. \ref{full}. As shown in Tab. \ref{tab1}, our SS-UIE demonstrates the best UIE performance on almost all full-reference metrics while requiring fewer Params and FLOPs (as shown in Fig. \ref{fig1}).
The visual comparisons shown in Fig. \ref{full} reveal that the enhancement results of our method are the closest to the reference image, which has high-fidelity local details and fewer color casts and artifacts. This demonstrates the effectiveness of using SS-block to capture the degradation degrees in different spatial regions and spectral bands, and performing degradation level-based adaptive UIE. It also proves that the parallel design can help the SS-block to achieve the interaction of spatial and spectral features while maintaining linear complexity, thereby obtaining both the spatial-wise and spectral-wise global receptive field.
\begin{table}[h]
	\centering
	\renewcommand\arraystretch{1.3}
	\tabcolsep=1.7pt
	\begin{tabular}{ccccccc|cc}
		\Xhline{1.5px}
		\rowcolor[HTML]{EFEFEF} 
		BL & SS2D & MCSS & SWSA & Serial & Parallel & FWL & PSNR & SSIM \\ \hline
		\checkmark  &      &      &      &        &          &     &22.67  &0.859      \\ \hline
		\checkmark  & \checkmark    &      &      &        &          &     &23.88      &0.873   \\ \hline
		\checkmark  &      & \checkmark    &      &        &          &     &24.47      &0.892    \\ \hline
		\checkmark  &      &      & \checkmark    &        &          &     &24.03      &0.879      \\ \hline
		\checkmark  &      & \checkmark    & \checkmark    & \checkmark      &          &     &26.87   &0.927      \\ \hline
		\checkmark  &      & \checkmark    & \checkmark    &        & \checkmark        &     &28.13    &0.938      \\ \hline
		\checkmark  &      & \checkmark    & \checkmark    &        & \checkmark        & \checkmark   &\textbf{28.87}  &\textbf{0.952}      \\ \Xhline{1.5px}
	\end{tabular}
	\caption{Break-down ablation study. Models are trained and tested on the LSUI dataset.}
	\label{tab2}
\end{table}

\textbf{Non-Reference Evaluation.} The U45 test set was utilized for the non-reference evaluation, in which statistical results and visual comparisons are shown in Tab. \ref{tab1} and Fig. \ref{non}. As shown in Tab. \ref{tab1}, our method achieved the highest scores on UIQM and UCIQE metrics, demonstrating the superior generalization performance of SS-UIE. Moreover, as shown in Fig. \ref{non}, the UIE performance of previous methods in high-frequency local details is unsatisfactory.  These methods either produce overly smooth results, sacrificing fine-grained textural details, or introduce undesirable speckled textures. In contrast, our SS-UIE can accurately reconstruct the high-frequency local details. This is because the FWL can reinforce the network’s attention in the regions with rich detail and improve the UIE quality in those regions.

\subsection{Ablation Study}

\begin{figure}[h] 
	\centering
	\includegraphics[width=1.0\linewidth]{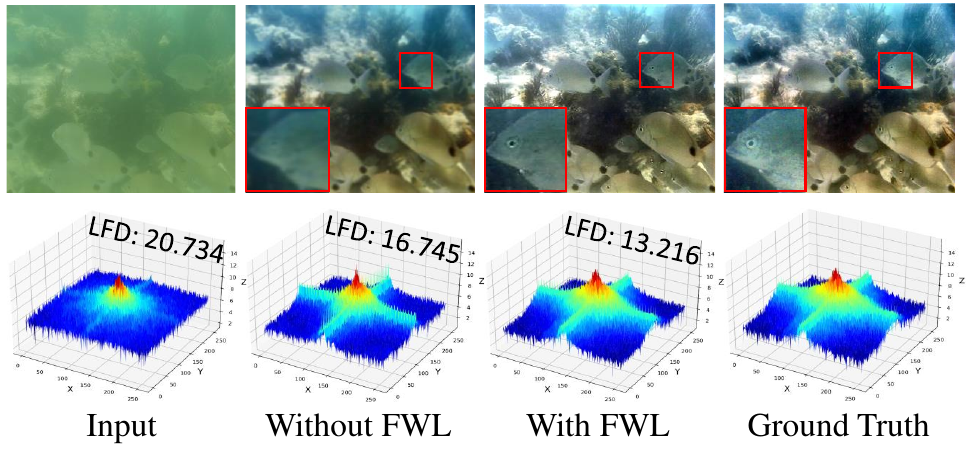}
	\caption{Frequency spectrum visualization with or without FWL. The metric LFD is used to measure the frequency similarity. The smaller the LFD, the closer the reconstructed image is to the ground truth. } 
	\label{fwl}
\end{figure}

To demonstrate the effectiveness of each component, we conducted a series of ablation studies on the LSUI dataset. We consider several factors including the SWSA and MCSS module in the SS-block, and the FWL function. The baseline (BL) model is derived by replacing our SS-block with the Swin transformer block \cite{liu2021swin}, and removing the FWL from SS-UIE. Serial and Parallel represent combining the SWSA and MCSS modules using serial and parallel design, respectively. Additionally, to validate the effectiveness of the multi-scale cycle selective scanning strategy in MCSS, we compared it with the original 2D Selective Scan (SS2D) in Vision-Mamba \cite{liu2024vmamba}.

\textbf{Effectiveness of SS-Blcok.} From Tab. \ref{tab2}, we observe that the removal of any component from the SS-block leads to a degradation in performance, which demonstrates the effectiveness of each component in the SS-block and the effectiveness of their combination. Moreover, the UIE performance of BL+MCSS is better than that of BL+ SS2D, which proves the effectiveness of the multi-scale cyclic selective scanning strategy. However, only using MCSS cannot capture the degradation levels between different spectral bands, which is why the reconstruction performance of BL+MCSS is not as good as BL+MCSS+SWSA. 
Furthermore, we can see that using a parallel design to combine MCSS and SWSA results in better UIE performance than using a serial design. Moreover, the computational and memory costs of parallel design (4.25M, 19.37G) are also fewer than serial design (5.78M, 28.95G). This is because in a serial design, the network processes spatial and spectral features in successive steps, which makes these two types of features less interweaved.
Conversely, the parallel design facilitates the interaction of spatial and spectral features while maintaining linear complexity, thereby enabling the SS-block to simultaneously capture spatial-wise and spectral-wise global receptive fields, model the degradation levels in different spatial regions and spectral bands, and perform degradation level-based adaptive UIE.

\textbf{Effectiveness of FWL.} 
We calculate the log frequency distance (LFD) as a frequency-level metric to evaluate the spectrum difference. The LFD has a logarithmic relationship with the frequency distance $d(u, v)$ in Eq. \ref{eq:17},
\begin{equation}
	\begin{aligned}
		{\rm F}_{LFD}=log\bigg[\frac{1}{HW}\bigg(\sum_{u=0}^{H-1}\sum_{v=0}^{W-1}|d(u,v)|\bigg)+1\bigg].
	\end{aligned}
	\label{eq:21}
\end{equation}
As shown in Fig. \ref{fwl}, we visualize the 3D-spectra reconstructed with or without the FWL and provide the corresponding LFD. It can be seen that the 3D-spectra optimized with our proposed FWL exhibits more accurate frequency reconstruction and lower LFD, aligning the frequency statistics more closely with the ground truth. On the contrary, the frequency 3D-spectra of the reconstructed image without FWL supervision loses a lot of frequency information, resulting in amplitude and phase distortions that manifest as blurred details and artifacts in the predicted image.
Additionally, as shown in Tab. \ref{tab2}, the inclusion of FWL leads to an increase in the PSNR and SSIM scores of SS-UIE. This is because fine-grained spectrum supervision can preserve more high-frequency information that is difficult to synthesize, thereby achieving higher UIE performance. Moreover, since FWL does not change the network structure, the improvements achieved by FWL will not bring any additional memory and computational costs during testing.

\section{Conclution}
In this paper, we explore how to achieve adaptive UIE based on the degradation levels in different spatial regions and spectral bands with linear complexity. To this end, we propose a novel adaptive dual-domain learning method, termed SS-UIE. Specifically, we first introduce the SWSA and MCSS modules and combine them in parallel to form a basic SS-block. MCSS and SWSA can obtain the spatial-wise and spectral-wise global receptive field with linear complexity , respectively. Parallel design allows the interaction of spatial and spectral features, and helps the SS-block to capture the degradation levels in different spatial regions and spectral bands, thereby achieving degradation level-based adaptive UIE. In addition, FWL is introduced for UIE to narrow the frequency-wise discrepancy. Dynamic frequency-level supervision forces the model to reconstruct fine-grained frequencies and improves the imaging quality in regions with high-frequency details. Extensive quantitative and qualitative experiments demonstrate that our SS-UIE outperforms other UIE methods while requiring cheaper computational and memory costs.

\section{Acknowledgments}
This work was supported by the National Natural Science Foundation of China (Nos. 61991451, 62322502, 62088101), and the Guangdong Province Key Laboratory of Intelligent Detection in Complex Environment of Aerospace, Land and Sea (2022KSYS016).

\bibliography{aaai25}

\newpage
\appendix

\section{A: Implementation details}

\subsection{A.1: Training details}
We use python and pytorch framework via NVIDIA RTX3090 on Ubuntu20 to implement the SS-UIE. The training set was enhanced by cropping, rotating and flipping the existing images. All images were adjusted to a fixed size (256*256) when input to the network, and the pixel value will be normalized to [0,1]. Adam optimization algorithm is utilized for a total of 600 epochs training with batchsize set as 10 and a learning rate of 0.0002. Besides, the learning rate decreased 60\% every 200 epochs.  The default values for $\beta_1$ and $\beta_2$ were set to 0.5 and 0.999, respectively. The weight decay was set to 0.00005.

\subsection{A.2: Detail structure of the SS-UIE network}
We designed the SS-UIE network based on the Spatial-Spectral block (SS-block). On this basis, our network further introduces dense connections among different SS-blocks, enabling long-distance dependency modeling and multi-level feature retrieval. The network consists of three modules, including the shallow feature extraction module, the deep feature fusion module, and the image reconstruction module.
\begin{table}[h]
	\centering
	\renewcommand\arraystretch{1.4}
	\begin{tabular}{c|c}
		\hline
		Shallow Feature Extraction Module & Output Size  \\ \hline
		BN(3)+Relu(3)+Conv(3,16,3,1,1)      & (256,256,16) \\
		BN(16)+Relu(16)+Conv(16,32,3,1,1)   & (256,256,32) \\
		Maxpool2d(2,2)                    & (128,128,32) \\
		BN(32)+Relu(32)+Conv(32,64,3,1,1)   & (128,128,64) \\
		Maxpool2d(2,2)                    & (64,64,64)   \\ \hline
	\end{tabular}
	\caption{Detail structure of the shallow feature extraction module. Conv(3,16,3,1,1) means that the number of input channels of the convolution layer is 3, the number of output channels is 16, the convolution kernel size is 3, the stride is 1, and the padding size is 1.}
	\label{tab1}
\end{table}

\textbf{Shallow Feature Extraction Module}. The detailed structure of the shallow feature extraction module is shown in Tab. \ref{tab1}. The shallow feature extraction module consists of three convolutional blocks and two max pooling layers for downsampling.  Each convolutional block consists of convolution, batch normalization, and activation layers.
The shallow feature extraction module is applied for preliminary visual processing, providing a simple way to map the input image space to a higher-dimensional feature space.

\textbf{Deep Feature Fusion Module}. The deep feature fusion module consists of 6 Densely Connected SS-Blocks (DCSSB) and a gated fusion block. The gated fusion block fuses the outputs of different DCSSB blocks with self-adaptive weights. The weight parameters $w_n$ are adaptively adjusted through backpropagation during network training. Such a module structure is conducive to deep mining of different levels of high dimensional features, which prevents losing long-term memory as the network deepens and enhances local details. 
Each DCSSB block consists of 4 densely connected SS-blocks and a convolutional block. Compared to the convolution blocks \cite{Li2019UIEB,Fabbri2018UGAN,Li2021UnderwaterIE,Islam2020FUnIE,li2017watergan,li2018cyclegan,uplavikar2019UIEDAL} and Transformer blocks \cite{Peng2023ushape,tang2023underwater,zhao2024wavelet} in conventional UIE networks , SS-block combines the MCSS and the SWSA module in parallel, which can obtain the spatial and spectral global receptive field with linear complexity, thereby capturing the degradation levels of different spatial regions and spectral bands, and achieving degradation level-based spatial-spectral dual-domain adaptive UIE.

\begin{table}[h]
	\centering
	\renewcommand\arraystretch{1.4}
	\begin{tabular}{c|c}
		\hline
		Image Reconstruction Module       & Output Size  \\ \hline
		Upsampling(scale factor=2)        & (128,128,64) \\
		BN(64)+Relu(64)+Conv(64,32,3,1,1)   & (128,128,32) \\
		Upsampling(scale factor=2)        & (256,256,32) \\
		BN(32)+Relu(32)+Conv(32,16,3,1,1)   & (256,256,16) \\
		BN(16)+Relu(16)+Conv(16,1,3,1,1)    & (256,256,3)   \\ \hline
	\end{tabular}
	\caption{Detail structure of the image reconstruction module.}
	\label{tab2}
\end{table}

\begin{figure*}[h] 
	\centering
	\includegraphics[width=1.0\linewidth]{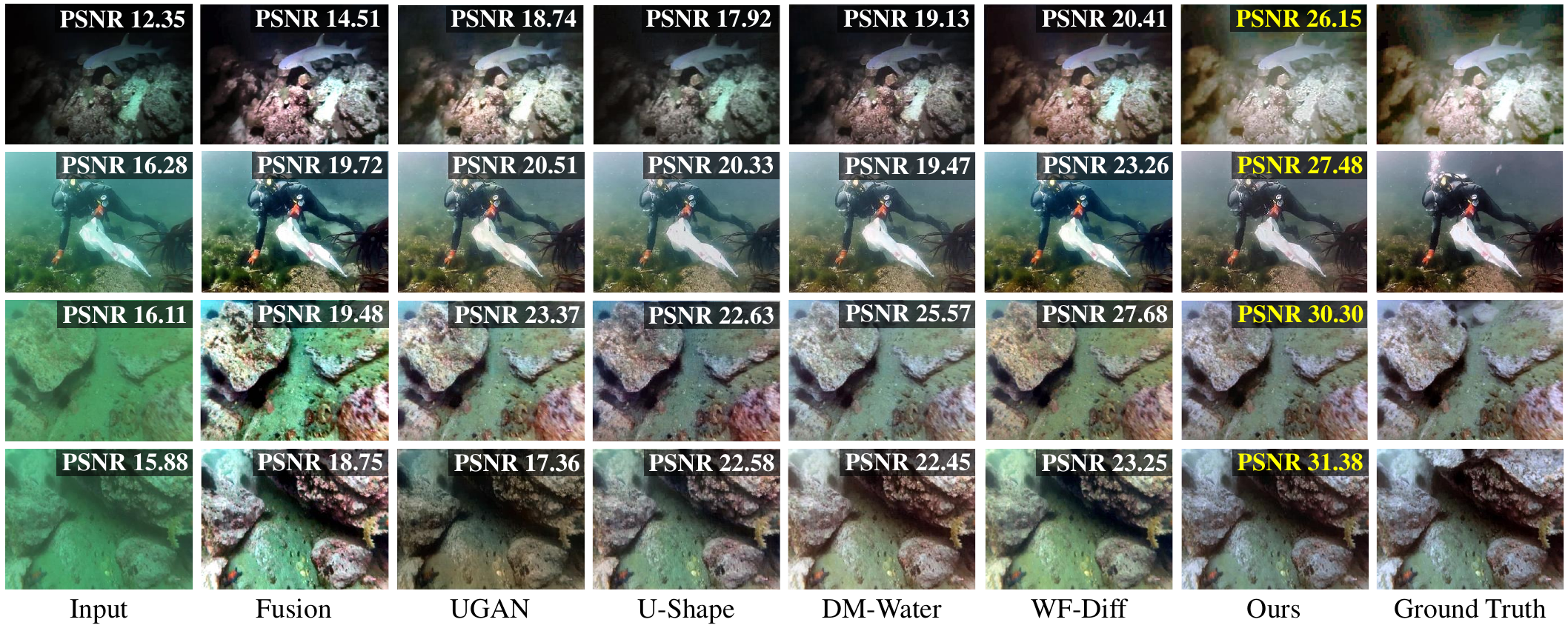}
	\vspace{-1.8em}
	\caption{Visual comparison of enhancement results sampled from the test set of LSUI \cite{Peng2023ushape} and UIEB \cite{Li2019UIEB} dataset. From left to right are raw underwater images, results of Fusion \cite{Ancuti2012fusion}, UGAN \cite{Fabbri2018UGAN}, U-Shape \cite{Peng2023ushape}, DM-Water \cite{tang2023underwater}, WF-Diff \cite{zhao2024wavelet}, our SS-UIE and the reference image (recognized as ground truth (GT)). The highest PSNR scores are marked in yellow. It can be seen that the enhancement results of our method are the closest to the ground truth. } 
	\label{full}
	\vspace{-1.0em}
\end{figure*}
\textbf{Image Reconstruction Module}. The detailed structure of the image reconstruction module is shown in Tab. \ref{tab2}. The image reconstruction module consists of three convolutional blocks and two upsampling layers. We retrieve high-quality underwater images by aggregating shallow features and multi-level deep fusion features. The shallow features $\mathbf{F_0}$ are mainly low-frequency image features, while the multi-level deep fusion features $\mathbf{F_{DF}}$ focus on recovering lost medium-frequency and high-frequency features. Benefiting from the long-term skip connections, the SS-UIE network can effectively transmit different-frequency information to final high-quality reconstruction.

\section{B: Supplementary experimental results}
In this section, we provide more visualized enhancement results of the full-reference and non-reference experiments. As shown in Fig. \ref{full}, the enhancement results of our method are the closest to the reference image, which has high-fidelity local details and fewer color casts and artifacts. This demonstrates the effectiveness of using SS-block to capture the degradation degrees in different spatial regions and spectral bands, and performing degradation level-based adaptive UIE. It also proves that the parallel design can help the SS-block to achieve the interaction of spatial and spectral features while maintaining linear complexity, thereby obtaining both the spatial-wise and spectral-wise global receptive field. 

Moreover, as shown in Fig. \ref{non}, the UIE performance of previous methods in high-frequency local details is unsatisfactory. They either produce overly smooth results, sacrificing fine-grained textural details, or introduce undesirable color artifacts and speckled textures. In contrast, our SS-UIE can accurately reconstruct the high-frequency local details. This is because the FWL can reinforce the network’s attention in the regions with rich detail and improve the UIE quality in those regions.

In addition, to verify the efficiency of our proposed SS-UIE and its potential for real-time applications for underwater exploration, we counted the inference time of all the learning-based UIE methods involved in the comparison (including Water-Net \cite{Li2019UIEB}, UGAN \cite{Fabbri2018UGAN}, U-Color \cite{Li2021UnderwaterIE}, U-Shape \cite{Peng2023ushape}, DM-Water \cite{tang2023underwater}, WF-Diff \cite{zhao2024wavelet} and our SS-UIE)  and presented the results in Tab. \ref{tab3}.

\begin{table*}[h]
	\centering
	\begin{tabular}{c|c|c|c|c|c|c|c}
		\hline
		Method         & Water-Net & UGAN & U-color & U-shape & DM-water & WF-Diff & Ours \\ \hline
		Inference time & 0.43      & 0.05 & 1.98    & 0.06    & 0.18     & 0.31    & 0.05 \\ \hline
	\end{tabular}
	\caption{The inference time of each learning-based UIE method. The data in the table is the inference time for images with a resolution of 256*256. The CPU used in the test is AMD R3700x and the GPU is NVIDIA RTX 3090.}
	\label{tab3}
\end{table*}
The results in Tab. \ref{tab3} shows that our SS-UIE achieves the best underwater image enhancement at a relatively fast speed of 20 FPS, meeting real-time requirements for underwater exploration.


\begin{figure*}[h] 
	\centering
	\includegraphics[width=1.0\linewidth]{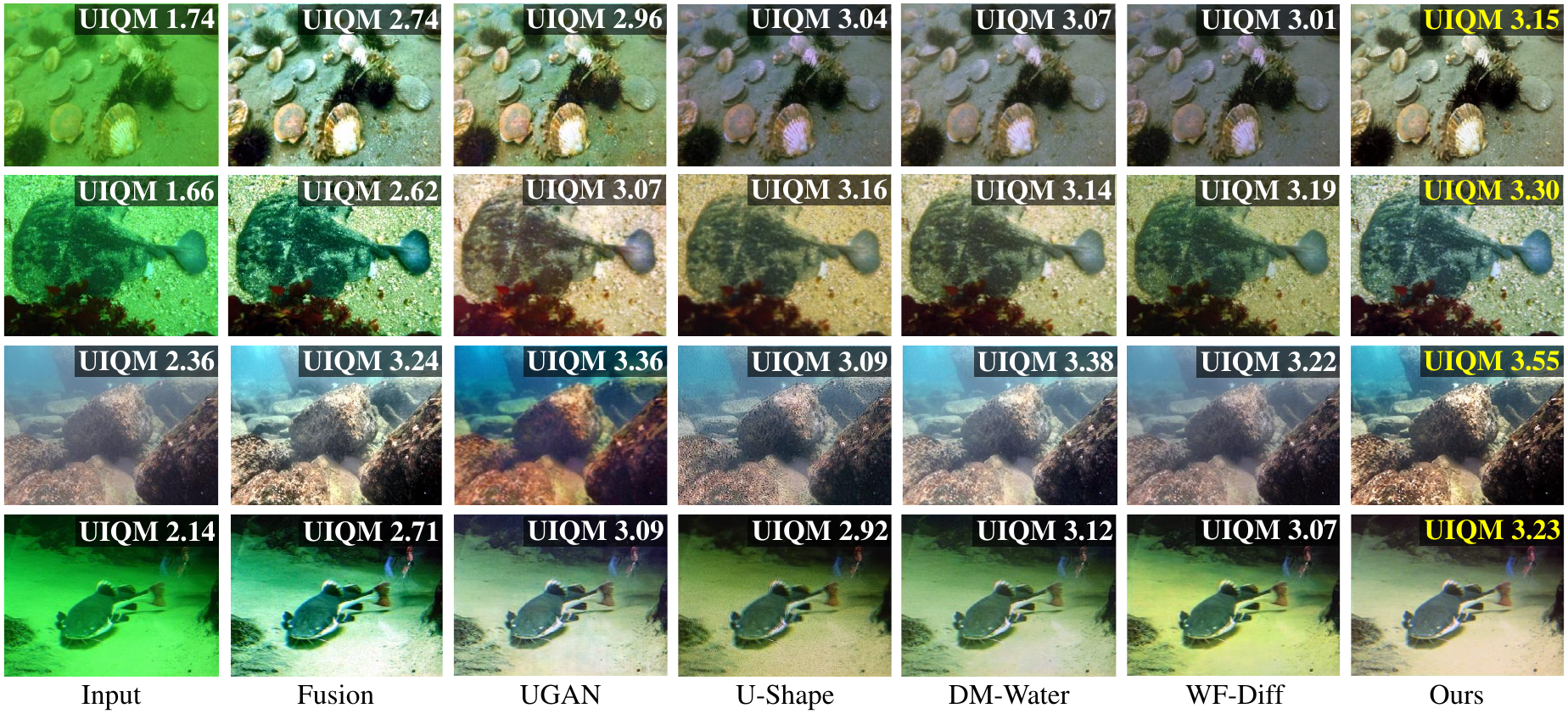}
	\vspace{-1.8em}
	\caption{Visual comparison of the non-reference evaluation sampled from the U45 \cite{li2019fusion} dataset. From left to right are raw underwater images, results of Fusion \cite{Ancuti2012fusion}, UGAN \cite{Fabbri2018UGAN}, U-Shape \cite{Peng2023ushape}, DM-Water \cite{tang2023underwater}, WF-Diff \cite{zhao2024wavelet} and our SS-UIE. Compared with existing methods, our method exhibits fewer color casts and artifacts, and recovers high-frequency local details better.} 
	\label{non}
	\vspace{-1.0em}
\end{figure*}

\section{C: Mathematical proof of DFT}
In this section, we derive the DFT from the standard Fourier transform (FT). The standard Fourier transform was originally designed for continuous signals. It converts continuous signals from the time domain to the frequency domain and can be regarded as an extension of the Fourier series. Specifically, the Fourier transform of the signal $x(t)$ is given by
\begin{equation}
	\begin{aligned}
		X(j\omega)=\int_{-\infty}^{\infty}x(t)e^{-j\omega t}dt:=\mathcal{F}[x(t)]
	\end{aligned},
	\label{eq:3}
\end{equation}
where $j$ is the imaginary unit. The inverse Fourier transform (IFT) has a similar form to the Fourier transform
\begin{equation}
	\begin{aligned}
		x(t)=\frac{1}{2\pi}\int_{-\infty}^{\infty}X(j\omega)e^{j\omega t}d\omega.
	\end{aligned}
	\label{eq:4}
\end{equation}

From Eq. \ref{eq:3} and Eq. \ref{eq:4} we can see the duality of FT between the time domain and the frequency domain. The duality indicates that the properties in the time domain always have their counterparts in the frequency domain. Fourier transform has a variety of properties. For example, the FT of a unit impulse function is
\begin{equation}
	\mathcal{F}[\delta(t)]=\int_{-\infty}^{\infty}\delta(t)e^{-j\omega t}dt=\int_{0-}^{0+}\delta(t)dt=1,
	\label{eq:5}
\end{equation}
and the time shifting property
\begin{equation}
	\begin{aligned}
		\mathcal{F}[\delta(t-t_0)]&=\int_{-\infty}^{\infty}x(t-t_0)e^{-j\omega t}dt \\
		& =e^{-j\omega t_0}\int_{-\infty}^{\infty}x(t)e^{-j\omega t}dt \\
		&=e^{-j\omega t_0}X(j\omega).
	\end{aligned}
	\label{eq:6}
\end{equation}

Since it is difficult for computers to directly process continuous signals, in practical applications, the general approach is to sample continuous signals to obtain discrete signal sequences. Sampling can be achieved using a sequence of unit impulse functions
\begin{equation}
	x_s(t)=x(t)\sum_{n=-\infty}^\infty\delta(t-nT_s)=\sum_{n=-\infty}^\infty x(nT_s)\delta(t-nT_s),
	\label{eq:7}
\end{equation}
where $T_s$ is the sampling interval. Taking the FT of the sampled signal $x_s(t)$ and applying Eq. \ref{eq:6} and Eq. \ref{eq:7}, we have
\begin{equation}
	X_s(j\omega)=\sum_{n=-\infty}^\infty x(nT_s)e^{-j\omega nT_s}.
	\label{eq:8}
\end{equation}
The above equation shows that $X_s(j\omega)$ is a periodic function with the fundamental period as $2\pi/Ts$. Actually, there is always a correspondence between the discrete signal in one domain and the periodic signal in the other domain. Generally, we prefer a normalized frequency $\omega\leftarrow\omega T_{s}$ such that the period of $X_s(j\omega)$ is exactly $2\pi$. We can further denote $x[n] = x(nT_s)$ as the sequence of discrete signal and derive the discrete-time Fourier transform (DTFT)
\begin{equation}
	X(e^{j\omega})=\sum_{n=-\infty}^{\infty}x[n]e^{-j\omega n}.
	\label{eq:9}
\end{equation}
Assuming that the discrete signal $x[n]$ has finite length N, then DTFT becomes
\begin{equation}
	X(e^{j\omega})=\sum_{n=0}^{N-1}x[n]e^{-j\omega n},
	\label{eq:10}
\end{equation}
without loss of generality, we assume that the nonzero terms lie in $[0, N-1]$. Note that the DTFT is a continuous function of $\omega$ and we can obtain a sequence of $X[k]$ by sampling $X(e^{j\omega})$ at frequencies $\omega_k=2\pi k/N$
\begin{equation}
	X[k]= X(e^{j\omega})\vert_{\omega=2\pi k/N}=\sum_{n=0}^{N-1}x[n]e^{-j(2\pi/N)kn}.
	\label{eq:11}
\end{equation}
This is exactly the formula for 1D DFT. DFT plays an important role in the area of digital signal processing and is a crucial component in our SF-block. 

Since DFT is a one-to-one transformation, given DFT $X[k]$, we can recover the original signal $x[n]$ by the inverse DFT (IDFT) as
\begin{equation}
	x[n]=\frac{1}{N}\sum_{k=0}^{N-1}X[k]e^{j(2\pi/N)kn}.
	\label{eq:13}
\end{equation}

Since DFT is conjugate symmetric, therefore for the real input $x[k]$, we can get $X[N-k] = X*[k]$. The reverse is true as well: if we perform IDFT to $X[k]$ which is conjugate symmetric, a real discrete signal can be recovered. This property implies that half of the DFT contains complete information about the frequency characteristics of $x[n]$. The fast Fourier transform (FFT) algorithms take advantage of the symmetry and periodicity properties of $W_{N}^{kn}$ and reduce the complexity to compute DFT from $O(N^2)$ to $O(NlogN)$. The inverse DFT (Eq. \ref{eq:13}), which has a similar form to the DFT, can also be computed efficiently using the inverse fast Fourier transform (IFFT).

The DFT described above can be extended to 2D signals, and the 2D DFT can be viewed as performing 1D DFT on the two dimensions alternatively. Given the 2D signal $X[m, n]$, $0 \leq m \leq M-1$, $0 \leq n \leq N-1$, the 2D DFT of $x[m, n]$ is given by
\begin{equation}
	\begin{aligned}
		X[u,v]&=\sum_{m=0}^{M-1}\sum_{n=0}^{N-1}x[m,n]e^{-j2\pi\left(\frac{um}{M}+\frac{vn}{N}\right)}.
	\end{aligned}
	\label{eq:14}
\end{equation}

Similar to 1D DFT, 2D DFT of real input $x[m,n]$ satisfied the conjugate symmetry property $X[M-u,N-v] = X*[u,v]$. The FFT algorithms can also be applied to 2D DFT to improve computational efficiency. In addition, FFT and IFFT operations are well-supported by hardware accelerators (like GPUs) through cuFFT and mkl-fft libraries, and there is a mature interface in Pytorch. This ensures that our SS-UIE can perform these computations fast, achieving stable and efficient training.

\end{document}